\documentclass[11pt,letterpaper]{article}
\usepackage{emnlp2016}
\usepackage{times}
\usepackage{latexsym}

\emnlpfinalcopy

\def\emnlppaperid{426}

\usepackage{url}
\usepackage{breakurl}
\usepackage{color}
\usepackage{tabularx}
\usepackage{ragged2e}
\newcolumntype{Y}{>{\RaggedRight\let\newline\\\arraybackslash\hspace{0pt}}X} 
\usepackage{multirow}
\usepackage{soul}
\usepackage{amsmath}

\usepackage{graphicx}
\usepackage{pdflscape}
\usepackage{afterpage}
\usepackage{xspace}
\usepackage{ifthen}
\usepackage{amssymb}
\usepackage{amsthm}

\usepackage{standalone}
\usepackage{tkz-berge}
\usepackage{tikz}
\usepackage{xcolor}
\usepackage{pgfplots}

\usepackage{diagbox}

\newcommand{\subscript}[1]{$_{[\text{#1}]}$}
\newcommand{\subscriptnum}[1]{$_{\text{#1}}$}
\newcommand{\squishlist}{
 \begin{list}{$\bullet$}
  { \setlength{\itemsep}{0pt}
     \setlength{\parsep}{3pt}
     \setlength{\topsep}{3pt}
     \setlength{\partopsep}{0pt}
     \setlength{\leftmargin}{1.5em}
     \setlength{\labelwidth}{1em}
     \setlength{\labelsep}{0.5em} } }

\newcounter{Lcount}
\newcommand{\squishlisttwo}{
\begin{list}{\arabic{Lcount}. }
{ \usecounter{Lcount}
\setlength{\itemsep}{0pt}
\setlength{\parsep}{0pt}
\setlength{\topsep}{0pt}
\setlength{\partopsep}{0pt}
\setlength{\leftmargin}{2em}
\setlength{\labelwidth}{1.5em}
\setlength{\labelsep}{0.5em} } }

    \newtheorem{theorem}{Theorem}[section]

    \renewenvironment{proof}[1][Proof Sketch]{\begin{trivlist}\pushQED{\qed}%
    \item[\hskip \labelsep {\bfseries #1}]}{\popQED\vspace{-1pt}\end{trivlist}}

\newcommand{\squishend}{
\end{list} }

\newcommand{\An}[1][]{$\mathbf{A}\ifthenelse{\equal{#1}{}}{}{_#1}$\xspace}
\newcommand{\En}[1][]{$\mathbf{E}\ifthenelse{\equal{#1}{}}{}{_#1}$\xspace}
\newcommand{\Tn}[1][]{$\mathbf{T}\ifthenelse{\equal{#1}{}}{}{_#1}$\xspace}
\newcommand{\Bn}[1][]{$\mathbf{B}\ifthenelse{\equal{#1}{}}{}{_#1}$\xspace}
\newcommand{\On}[1][]{$\mathbf{O}\ifthenelse{\equal{#1}{}}{}{_#1}$\xspace}
\newcommand{\Xn}{$\mathbf{X}$\xspace}
\newcommand{\Shared}{\textsc{Shared}\xspace}
\newcommand{\Split}{\textsc{Split}\xspace}

\title{Learning to Recognize Discontiguous Entities}

\author{Aldrian Obaja Muis \qquad Wei Lu \\
  Singapore University of Technology and Design \\
  {\tt \{aldrian\_muis,luwei\}@sutd.edu.sg}}

\date{}

\begin{document}
\maketitle
\begin{abstract}
This paper focuses on the study of recognizing discontiguous entities.
Motivated by a previous work, we propose to use a novel hypergraph representation to jointly encode discontiguous entities of unbounded length, which can overlap with one another.
To compare with existing approaches, we first formally introduce the notion of {\em model ambiguity}, which defines the difficulty level of interpreting the outputs of a model, and then formally analyze the theoretical advantages of our model over previous existing approaches based on linear-chain CRFs.
Our empirical results also show that our model is able to achieve significantly better results when evaluated on standard data with many discontiguous entities.
\end{abstract}

\section{Introduction}

Building effective automatic named entity recognition (NER) systems that is capable of extracting useful semantic shallow information from texts has been one of the most important tasks in the field of natural language processing. 
An effective NER system can typically play an important role in certain downstream NLP tasks such as relation extraction, event extraction, and knowledge base construction \cite{Hasegawa2004b,Al-Rfou2012}.

\begin{figure}[t!]
\centering
\footnotesize
{\def\arraystretch{0.9}\tabcolsep=2pt
\begin{tabularx}{0.98\columnwidth}{|Y|}
\hline
\\
{\textnormal{\ \ \ } EGD showed [{\bf {\em hiatal hernia}}]\subscriptnum{1} and vertical [{\bf{\em laceration}}]\subscriptnum{2}} 

{\textnormal{\ \ \ }  in distal [{\bf{\em esophagus}}]\subscriptnum{2} with [{\bf {\em blood in}} [{\bf {\em stomach}}]\subscriptnum{4}]\subscriptnum{3} and}

{\textnormal{\ \ \ } overlying [{\bf{\em lac}}]\subscriptnum{4}.}\\\\\hline
\end{tabularx}
}
\vspace{1mm}
\caption{Discontiguous  entities in a medical domain. Words annotated with the same index are part of the same entity. Note that entity 3 and entity 4 overlap with one another.}
\label{ct-sample-1}
\vspace{-1mm}
\end{figure}

Most traditional NER systems are capable of extracting entities\footnote{Or sometimes {\em mentions} are considered, which can be named, nominal or pronominal references to entities \cite{Florian:2004uv}. In this paper we use ``mentions'' and ``entities'' interchangeably.} as short spans of texts.
Two basic assumptions are typically made when extracting entities: 1) entities do not overlap with one another, and 2) each entity consists of a contiguous sequence of words.
These assumptions allow the task to be modeled as a sequence labeling task, for which many existing models are readily available, such as linear-chain CRFs \cite{McCallum2003}. 

While the above two assumptions are valid for most cases, they are not always true.
For example, in the entity {\em University of New Hampshire} of type {\em ORG} there exists another entity {\em New Hampshire} of type {\em LOC}.
This violates the first assumption above,
yet it is crucial to extract both entities for subsequent tasks such as relation extraction and knowledge base construction.
Researchers therefore have proposed to tackle the above issues in NER using more sophisticated models \cite{Finkel2009,Lu2015}.
Such efforts still largely rely on the second assumption. 

Unfortunately, the second assumption is also not always true in practice.
There are also cases where the entities are composed of multiple discontiguous sequences of words, such as in disorder mention recognition in clinical texts \cite{Pradhan2014a}, where the entities (disorder mentions in this case) may be discontiguous.
Consider the example shown in Figure \ref{ct-sample-1}.
In this example there are four entities, the first one, {\bf {\em hiatal hernia}}, is a conventional contiguous entity. 
The second one, {\bf{\em laceration ... esophagus}}, is a discontiguous entity, consisting of two parts. 
The third and fourth ones, {\bf{\em blood in stomach}} and {\bf{\em stomach ... lac}} (for {\em stomach laceration}), are overlapping with each other, with the fourth being discontiguous at the same time.

For such discontiguous entities which can potentially overlap with other entities in complex manners, existing approaches such as those based on simple sequence tagging models have difficulties handling them accurately. 
This stems from the fact that there is a very large number of possible entity combinations in a sentence when the entities can be discontiguous and overlapping.

Motivated by this, in this paper we propose a novel model that can better represent both contiguous and  discontiguous entities which can overlap with one another. Our major contributions can be summarized as follows:
\squishlist
\item We propose a novel model that is able to represent both contiguous and discontiguous entities. 
\item Theoretically, we introduce the notion of {\em model ambiguity} for quantifying the ambiguity of different NER models that can handle discontiguous entities. 
We present a study and make comparisons about different models' ambiguity under this theoretical framework.
\item Empirically, we demonstrate that our model can significantly outperform conventional approaches designed for handling discontiguous entities on data which contains many discontiguous entities.
\squishend

\section{Related Work}

Learning to recognize named entities is a popular task in the field of natural language processing.
A survey by Nadeau \shortcite{Nadeau2007} lists several approaches in NER, including Hidden Markov Models (HMM) \cite{Bikel1997}, Decision Trees \cite{Sekine1998}, Maximum Entropy Models \cite{Borthwick1998}, Support Vector Machines (SVM) \cite{Asahara2003}, and also semi-supervised and unsupervised approaches. Ratinov \shortcite{Ratinov2009} utilizes averaged perceptron to solve this problem and also focused on four key design decisions, achieving state-of-the-art in MUC-7 dataset.
These approaches work on standard texts, such as news articles, and the entities to be recognized are defined to be contiguous and non-overlapping.

Noticing that many named entities contain other named entities inside them, Finkel and Manning \shortcite{Finkel2009} proposed a model that is capable of extracting nested named entities by representing the sentence as a constituency parse tree, with named entities as phrases. 
As a parsing-based model, the approach has a time complexity that is cubic in the number of words in the sentence.

Recently, Lu and Roth \shortcite{Lu2015} proposed a model that can represent overlapping entities.
In addition to supporting nested entities, theoretically this model can also represent overlapping entities where neither is nested in another.
The model represents each sentence as a hypergraph with nodes indicating entity types and boundaries.
Compared to the previous model, this model has a lower time complexity, which is linear in the number of words in the sentence.

All the above models focus on NER in conventional texts, where the assumption of contiguous entities is valid.
In the past few years, there is a growing body of works on recognizing disorder mentions in clinical text. 
These disorder mentions may be discontiguous and also overlapping.
%
To tackle such an issue, a research group from University of Texas Health Science Center at Houston \cite{Tang2013,Zhang2014,Xu2015} first utilized a conventional linear-chain CRF to recognize disorder mention parts by extending the standard BIO ({\em Begin, Inside, Outside}) format, and next did some postprocessing to combine different components.
Though effective, as we will see later, such a model comes with some drawbacks.
Nevertheless, their work motivated us to perform further analysis on this issue and propose a novel model specifically designed for discontiguous entity extraction.

\section{Models}

\subsection{Linear-chain CRF Model}

Before we present our approach, we would like to spend some time to discuss a simple approach based on linear-chain CRFs \cite{Lafferty:2001ty}. This approach is primarily based on the system by Tang et al. \shortcite{Tang2013},
and this will be the baseline system that we will make comparison with in later sections.

The problem is regarded as a sequence prediction task, where each word is assigned a label similar to BIO format often used for NER. We used the encoding used by Tang et al. \shortcite{Tang2013}, which uses 7 tags to handle entities that can be discontiguous and overlapping. 
Specifically, we used $\textbf{B}$, $\textbf{I}$, $\textbf{O}$, $\textbf{BD}$, $\textbf{ID}$, $\textbf{BH}$, and $\textbf{IH}$ to denote $\textbf{B}$eginning of entity, $\textbf{I}$nside entity, $\textbf{O}$utside of entity, $\textbf{B}$eginning of $\textbf{D}$iscontiguous entity, $\textbf{I}$nside of $\textbf{D}$iscontiguous entity, $\textbf{B}$eginning of $\textbf{H}$ead, and $\textbf{I}$nside of $\textbf{H}$ead. To encode a sentence in this format, first we identify the contiguous word sequences which are parts of multiple entities. We call these \textit{head components} and we label each word inside each component with $\textbf{BH}$ (for the first word in {\em each component}) or $\textbf{IH}$.
Then we find contiguous word sequences which are parts of a discontiguous entity, which we call the \textit{body components}. Words inside those components which have not been labeled are labeled with $\textbf{BD}$ (for the first word in {\em each component}) or $\textbf{ID}$.
Finally, words that are parts of a contiguous entity are called \textit{contiguous component}, and, if they have not been labeled, are labeled as $\textbf{B}$ (for the first word in {\em each component}) or $\textbf{I}$.

\begin{figure}[t]
\footnotesize
{\def\arraystretch{0.9}\tabcolsep=2pt
\begin{tabularx}{\columnwidth}{|@{\hskip 2pt}Y@{\hskip 2pt}|}\hline
\\
{\ \ \ EGD showed {\bf{\em hiatal}}\subscript{\bf B} {\bf{\em hernia}}\subscript{\bf I} and vertical {\bf{\em laceration}}\subscript{\bf BD} }

{\ \ \ in distal {\bf{\em esophagus}}\subscript{\bf BD} with {\bf{\em blood}}\subscript{\bf B} {\bf{\em in}}\subscript{\bf I} {\bf{\em stomach}}\subscript{\bf BH} and} 

{\ \ \  overlying {\bf{\em lac}}\subscript{\bf BD}.}
\\
\\\hline
\\
{\ \ \ \textbf{\em Infarctions}\subscript{\bf BH} either \textbf{\em water}\subscript{\bf BD} \textbf{\em shed}\subscript{\bf ID} or \textbf{\em embolic}\subscript{\bf BD}}
\\%
\\\hline%
\end{tabularx}
}
\vspace{0mm}
\caption{Entity encoding in the linear-chain model. {\bf Top}: for the example in Fig \ref{ct-sample-1}. {\bf Bottom}: for the second example in Fig \ref{ct-sample-2}. The {\bf O} labels are not shown. Misspellings are from the dataset.}
\vspace{-4mm}
\label{lcrf-sample-1}
\end{figure}

This encoding is lossy, since the information on which parts constitute the same entity is lost. The top example in Figure \ref{lcrf-sample-1} is the encoding of the example shown in Figure \ref{ct-sample-1}. During decoding, based on the labels only it is not entirely clear whether ``\textbf{\em laceration}'' should be combined with ``\textbf{\em esophagus}'' or with ``\textbf{\em stomach}'' to form a single mention. For the bottom example, we cannot deduce that ``\textbf{\em Infarctions}'' alone is a mention, since there is no difference in the encoding of a sentence with only two mentions \{``\textbf{\em Infarctions \dots water shed}'', ``\textbf{\em Infarctions \dots embolic}''\} or having three mentions with ``\textbf{\em Infarctions}'' as another mention, since in both cases, the word ``\textbf{\em Infarctions}'' is labeled with \textbf{BH}.

Also, it should be noted that some of the label sequences are not valid. For example, a sentence in which there is only one word labeled as \textbf{BD} is invalid, since a discontiguous entity requires at least two words to be labeled as \textbf{BD} or \textbf{BH}. This is, however, a possible output from the linear CRF model, due to the Markov assumption inherent in linear CRF models. Later we see that our models do not have this problem.

\subsection{Our Model}
\label{our-model}

Linear-chain CRF models are limited in their representational power when handling complex entities, especially when they can be discontiguous and can overlap with one another.
While recent models have been proposed to effectively handle overlapping entities, how to effectively handle discontiguous entities remains a research question to be answered.
Motivated by previous efforts on handling overlapping entities \cite{Lu2015}, in this work we propose a model based on hypergraphs that can better represent entities that can be discontiguous and at the same time be overlapping with each other.

Unlike the previous work \cite{Lu2015}, we establish a novel theoretical framework to formally quantify the ambiguity of our hypergraph-based models and justify their effectiveness by making comparisons with the linear-chain CRF approach.

Now let us introduce our novel hypergraph representation.
A hypergraph can be used to represent entities of different types and their combinations in a given sentence.
Specifically, a hypergraph is constructed as follows. For the word at position $k$, we have the following nodes:
\squishlist
\item \An[{}^k]: this node represents all entities that begin with the current or a future word (to the right of the current word).
\item \En[{}^k]: this node represents all entities that begin with the current word.
\item \Tn[{t}^k]: this node represents entities of certain specific type $t$ that begin with the current word. There is one \Tn[{t}^k] for each different type.
\item \Bn[{t,i}^k]: this node indicates that the current word is part of the $i$-th component of an entity of type $t$.
\item \On[{t,i}^k]: this node indicates that the current word appears in between $(i$-$1)$-th and $i$-th components of an entity of type $t$.
\squishend

There is also a special leaf node, \Xn-node, which indicates the end ({\em i.e.}, right boundary) of an entity.

The nodes are connected by directed hyperedges, which for the purpose of explaining our models are defined as those edges that connect one node, called the parent node, to one or more child nodes. For ease of notation, in the rest of this paper we use \textit{edge} to refer to directed hyperedge.


\vspace{-1mm}

\paragraph{The edges} Each \An[{}^k] is a parent to \En[{}^k] and \An[{}^{k+1}], encoding the fact that the set of all entities at position $k$ is the union of the set of entities starting \textbf{e}xactly at current position (\En[{}^k]) with the set of entities starting at or \textbf{a}fter position $k+1$ (\An[{}^{k+1}]).

Each \En[{}^k] is a parent to \Tn[{1}^k], $\dots$, \Tn[{T}^k] , where $T$ is the total number of possible types that we consider.
Each \Tn[{t}^k] has two edges where it serves as a parent, within one it is parent to \Bn[{t,0}^k] and within another it is to \Xn. These edges encode the fact that at position $k$, either there is an entity of type $t$ that begins with the current word (to \Bn[{t,0}^k]), or there is no entity of type $t$ that begins with the current word (to \Xn{}).

In the full hypergraph, each \Bn[{t,i}^k] is a parent to \Bn[{t,i}^{k+1}] (encoding the fact that the next word also belongs to the same component of the same entity), to \On[{t,i+1}^{k+1}] (encoding the fact that this word is part of a discontiguous entity, and the next word is the first word separating current component and the next component), and to \Xn (representing that the entity ends at this word). Also there are edges with all possible combinations of \Bn[{t,i}^{k+1}], \On[{t,i+1}^{k+1}], and \Xn as the child nodes, representing overlapping entities. For example, the edge \Bn[{t,i}^k]$\to ($\Bn[{t,i}^{k+1}],\Xn$)$ denotes that there is an entity which continues to the next word (the edge to \Bn[{t,i}^{k+1}]), while there is another entity ending at $k$-th word (the edge to \Xn). In total there are 7 edges in which \Bn[{t,i}^k] is a parent, which are:
\squishlist
\item \Bn[{t,i}^k] $\rightarrow ($\Xn\!$)$
\item \Bn[{t,i}^k] $\rightarrow ($\On[{t,i+1}^{k+1}]\!$)$
\item \Bn[{t,i}^k] $\rightarrow ($\On[{t,i+1}^{k+1}]$,$ \Xn\!$)$
\item \Bn[{t,i}^k] $\rightarrow ($\Bn[{t,i}^{k+1}]\!$)$
\item \Bn[{t,i}^k] $\rightarrow ($\Bn[{t,i}^{k+1}]$,$ \Xn\!$)$
\item \Bn[{t,i}^k] $\rightarrow ($\Bn[{t,i}^{k+1}]$,$ \On[{t,i+1}^{k+1}]\!$)$
\item \Bn[{t,i}^k] $\rightarrow ($\Bn[{t,i}^{k+1}]$,$ \On[{t,i+1}^{k+1}]$,$ \Xn\!$)$
\end{list}

Analogously, \On[{t,i}^k] has three edges that connect to \On[{t,i}^{k+1}], \Bn[{t,i+1}^{k+1}], and both. Note that \On[{t,i}^k] is not a parent to \Xn by definition.

\begin{figure}
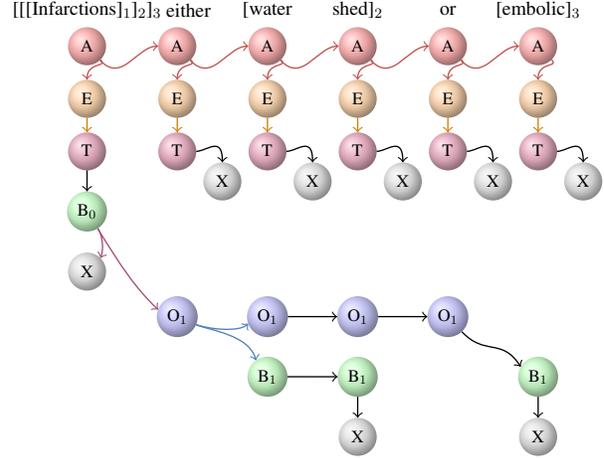

\centering
\includestandalone[width=0.99\columnwidth]{emnlp-model}
\vspace{-6mm}
\caption{The hypergraph for \textsc{Shared} model for the second example in Figure \ref{ct-sample-2}. The type information in \Tn, \Bn, and \On-nodes is not shown. The \Xn{}-node is drawn multiple times for better visualization.}
\label{hypergraph-1}
\vspace{-4mm}
\end{figure}

During testing, the model will predict a \textbf{subgraph} which will result in the predicted entities after decoding. We call this subgraph representing certain entity combination \textbf{\em entity-encoded hypergraph}.

For example, Figure \ref{hypergraph-1} shows the entity-encoded hypergraph of our model encoding the three mentions in the second example in Figure \ref{ct-sample-2}. The edge from the \Tn-node for the first word to the \Bn-node for the first word shows that there is at least one entity starting with this word. The three places where an \Xn-node is connected to a \Bn-node show the end of the three entities. Note that this hypergraph clearly shows the presence of the three mentions without ambiguity, unlike a linear-chain encoding of this example where it cannot be inferred that ``\textbf{\em Infarctions}'' alone is a mention, as discussed previously.
In this paper, we set the maximum number of components to be 3 since the dataset does not contain any mention with more than 3 components.

Also note that this model supports discontiguous and overlapping mentions of different types since each type has its own set of \On{}-nodes and \Bn{}-nodes, unlike the linear-chain model, which supports only overlapping mentions of the same type.


We also experimented with a variant of this model, where we split the \Tn-nodes, \Bn-nodes, and \On-nodes further according to the number of components. We split \Bn[{t,i}^k] into \Bn[{t,i,j}^k], $i=1\ldots j, j=1\ldots3$ which represents that the word is part of the $i$-th component of a mention with total $j$ components. Similarly we split \On[{t,i}^k] into \On[{t,i,j}^k] and \Tn[{t}^k] into \Tn[{t,j}^k]. We call the original version \Shared model, and this variant \Split model. The motivation for this variant is that the majority of overlaps in the data are between discontiguous and contiguous entities, and so splitting the two cases -- one component (contiguous) and more (discontiguous) -- will reduce ambiguity for those cases.

These models are still ambiguous to some degree, for example when an \On-node has two child nodes and two parents, we cannot decide which of the parent node is paired with which child node.
However, in this paper we argue that:

\squishlist
\item This model is less ambiguous compared to the linear-chain model, as we will show later theoretically and empirically.
\item Every output of our model is a valid prediction, unlike the linear-chain model since this model will always produce a valid path from \Tn-nodes to the \Xn-nodes representing some entities.
\squishend

We will also show through experiments that our models
 can encode the entities more accurately.

\subsection{Interpreting Output Structures}
\label{decoding}

Both the linear-chain CRF model and our models are still ambiguous to some degree, so we need to handle the ambiguity in interpreting the output structures into entities.
For all models, we define two general heuristics: \textsc{Enough} and \textsc{All}. The \textsc{Enough} heuristic handles ambiguity by trying to produce a minimal set of entities which encodes to the one produced by the model, while \textsc{All} heuristic handles ambiguity by producing the union of all possible entity combinations that encode to the one produced by the model.
For more details on how these heuristics are implemented for each model, please refer to the supplementary material.

\subsection{Training}
\label{sec:training}

For both models, the training follows a log-linear formulation, by maximizing the loglikelihood of the training data $\mathcal{D}$:

\begin{small}
\vspace{-7mm}
\begin{eqnarray}
\mathcal{L}(\mathcal{D}) =
\!\!\!\!\!\!
\sum_{(\mathbf{x},\mathbf{y})\in\mathcal{D}}
\!\!
\left[\sum_{e\in \mathcal{E}(\mathbf{x},\mathbf{y})}
\!\!\!\!
\left[\mathbf{w}^T\mathbf{f}(e)\right]
- \log Z_{\mathbf{w}}(\mathbf{x})\vphantom{\sum_{e\in\mathcal{E}(\mathbf{x}}}\right]
\!\!
-
\!
\lambda ||\mathbf{w}||^2
\nonumber
\end{eqnarray}
\end{small}
Here $(\mathbf{x}, \mathbf{y})$ is a training instance consisting of the sentence $\mathbf{x}$ and the entity-encoded hypergraph $\mathbf{y}\in\mathcal{Y}$ where $\mathcal{Y}$ is the set of all possible mention-encoded hypergraphs.
The vector $\mathbf{w}$ consists of feature weights, which are the model parameters to be learned.
The set $\mathcal{E}(\mathbf{x},\mathbf{y})$ consists of all edges present in the entity-encoded hypergraph $\mathbf{y}$ for input $\mathbf{x}$.
The function $\mathbf{f}(e)$ returns the features defined over the edge $e$, $Z_{\mathbf{w}}(\mathbf{x})$ is the normalization term which gives the sum of scores over all possible entity-encoded hypergraphs in $\mathcal{Y}$ that is relevant to the input $\mathbf{x}$, and finally $\lambda$ is the $\ell_2$-regularization parameter.



\section{Model Ambiguity}
\label{ambiguity}

The main aim of this paper is to assess how well each model can represent the discontiguous entities, even in the presence of overlapping entities.

In this section, we will theoretically compare the models' ambiguity, which is defined as the average number of mention combinations that map to the same encoding in a model. Now, to compare two models, instead of calculating the ambiguity directly, we can calculate the \textbf{\em relative ambiguity} between the two models directly by comparing the number of \textit{canonical encodings} in the two models.

A canonical encoding is a fixed, selected representation of a particular set of mentions in a sentence, among (possibly) several alternative representations. Several alternatives may be present due to the ambiguity of the encoding-decoding process and also since the output of the model is not restricted to a specific rule. For example, for the text ``John Smith'', a model trained in BIO format might output ``B-PER I-PER'' or ``I-PER I-PER'', and both will still mean that ``John Smith'' is a person, although the ``correct'' encoding would of course be ``B-PER I-PER'', which is selected as the canonical encoding. Intuitively, a canonical encoding is a formal way to say that we only consider the ``correct'' encodings.

A model with larger number of canonical encodings will, on average, have less ambiguity compared to the one with smaller number of canonical encodings. Subsequently, a model with less ambiguity will be more precise in predicting entities.

Let $\mathcal{M}_{\textsc{Li}}(n), \mathcal{M}_{\textsc{Sh}}(n), \mathcal{M}_{\textsc{Sp}}(n)$ denote the number of canonical encodings of the linear-chain, \Shared, and \Split model, respectively, for a sentence with $n$ words. Then we formally define the relative ambiguity of model $M_1$ over model $M_2$, $\mathcal{A}_r(M_1, M_2)$, as follows:
\begin{equation}
\mathcal{A}_r(M_1, M_2) =  \lim_{n\to\infty} \frac{\log\sum_{i=1}^n\mathcal{M}_{M_2}(i)}{\log\sum_{i=1}^n\mathcal{M}_{M_1}(i)}
\end{equation}

$\mathcal{A}_r(M_1, M_2) > 1$ means model $M_1$ is more ambiguous than $M_2$.
Now, we claim the following:
\begin{theorem}
\label{th:more-ambiguous}
$\mathcal{A}_r(\textsc{Li}, \textsc{Sh}) > 1$
\end{theorem}

We provide a proof sketch below.
Due to space limitation, we cannot provide the full dynamic programming calculation. We refer the reader to the supplementary material for the details.
\begin{proof}
The number of canonical encodings in the linear-chain model is less than $7^n$ since there are 7 possible tags for each of the $n$ words and not all of the $7^n$ tag sequences are canonical encodings. So we have $\mathcal{M}_{\textsc{Li}}(n) < 7^n$ and thus we can derive $\log \sum_{i=1}^n\mathcal{M}_{\textsc{Li}}(i) < 3n\log 2$.

For our models, by employing some dynamic programming adapted from the inside algorithm \cite{Baker1979}, we can calculate the growth order of the number of canonical encodings for \Shared model to arrive at a conclusion that $\forall n>n_0,\ \sum_{i=1}^n\mathcal{M}_{\textsc{Sh}}(i) > C\cdot2^{10n}$ for some constants $n_0, C$. Then we have:

\vspace{-3mm}

\begin{equation}
\mathcal{A}_r(\textsc{Li},\textsc{Sh})\!\geq\!\lim_{n\to\infty}\!\!\frac{\log C\!+\!10n\log 2}{3n\log 2}\!=\!\frac{10}{3}\!>\!1 \qedhere
\end{equation}
\end{proof}

\vspace{-3mm}

Theorem \ref{th:more-ambiguous} says that the linear-chain model is more ambiguous compared to our \Shared model. Similarly, we can also establish $\mathcal{A}_r(\textsc{Sh},\textsc{Sp}) > 1$. Later we also see this empirically from experiments.

\section{Experiments}

\subsection{Data}

To allow us to conduct experiments to empirically assess different models' capability in handling entities that can be discontiguous and can potentially overlap with one another, we need a text corpus annotated with entities which can be discontiguous and overlapping with other entities.
We found the largest of such corpus to be the dataset from the task to recognize disorder mentions in clinical text, initially organized by ShARe/CLEF eHealth Evaluation Lab (SHEL) in 2013 \cite{Suominen2013} and continued in SemEval-2014 \cite{Pradhan2014}.

The definition of the task is to recognize mentions of concepts that belong to the Unified Medical Language System (UMLS) semantic group \textit{disorders} from a set of clinical texts.
Each text has been annotated with a list of disorder mentions by two professional coders trained for this task, followed by an open adjudication step \cite{Suominen2013}.


Unfortunately, even in this dataset, only 8.95\% of the mentions are discontiguous.
Working directly on such data would prevent us from understanding the true effectiveness of different models when handling entities which can be discontiguous and overlapping.
In order to truly understand how different models behave on data with discontiguous entities, we consider a subset of the data where we consider those sentences which contain at least one discontiguous entity. We call the resulting subset the ``Discontiguous'' subset of the ``Original'' dataset. Later we will also still use the training data of the ``Original'' dataset in the experiments.

Note that this ``Discontiguous'' subset still contains contiguous entities since a sentence usually contains more than one entity. The subset is a balanced dataset with 53.61\% of the entities being discontiguous and the rest contiguous. We then split this dataset into training, development, and test set, according to the split given in SemEval 2014 setting (henceforth \textsc{large} dataset). To see the impact of dataset size, we also experiment on a subset of the \textsc{large} dataset, following the SHEL 2013 setting, with the development set in the \textsc{large} dataset used as test set (henceforth \textsc{small} dataset). The training and development set of the \textsc{small} dataset comes from a random 80\% (Tr80) and 20\% (Tr20) split of the training set in \textsc{large} dataset.

The statistics of the datasets, including the number of overlaps between the entities in the ``All'' column, are shown in Table \ref{data-stats}.

We note that this dataset only contains one type of entity. In later experiments, in order to evaluate the models on multiple types, we create another dataset where we split the entities based on the entity-level semantic category. This information is available for some entities through the Concept Unique Identifier (CUI) annotation in the data. In total we have three types: two types (type \texttt{A} and \texttt{B}) based on the semantic category, and one type (type \texttt{N}) for those entities having no semantic category information\footnote{It is tempting to just ignore these entities since the \texttt{N} type does not convey any specific information about the entities in it. However, due to the dataset size, excluding this type will lead to very small number of interactions between types. So we decided to keep this type}. See the supplementary material for more details. The number of overlaps between different types is shown in the ``Diff'' column in Table \ref{data-stats}. Except for a handful overlaps in development set, all overlaps involve at least one discontiguous entity. Our main result will still be based on the dataset with one type of entity.


\begin{table}
\footnotesize
\centering
{\def\arraystretch{1.15}\tabcolsep=2pt
\begin{tabular}{l|r|r|r|r|r|r|r}
\multirow{2}{*}{Split} & \multirow{2}{*}{\#Sentences} & \multicolumn{4}{c|}{Number of mentions} & \multicolumn{2}{c}{\#Overlaps}\\
& & \multicolumn{1}{c|}{1 part} & \multicolumn{1}{c|}{2 parts} & \multicolumn{1}{c|}{3 parts} & \multicolumn{1}{c|}{Total} & All & Diff \\\hline
Train & 534 & 544 & 607 & 44 & 1,195 & 205 & 58\\
- Tr80 & 416 & 448 & 476 & 33 & 957 & 164 & 48 \\
- Tr20 & 118 & 96 & 131 & 11 & 238 & 41 & 10 \\
Dev & 303 & 357 & 421 & 18 & 796 & 240 & 28\\
Test & 430 & 584 & 610 & 16 & 1,210 & 327 & 61\\
\end{tabular}
}
\vspace{0mm}
\caption{The statistics of the data. Tr80 and Tr20 refers to the 80\% and 20\% partitions of the full training data.}
\label{data-stats}
\vspace{-3mm}
\end{table}

\begin{figure}[ht]
\footnotesize
\vspace{-1mm}
\centering
{\def\arraystretch{0.9}\tabcolsep=2pt
\begin{tabularx}{\columnwidth}{|Y|}
\hline
\\
{\ \ \ The patient had blood in his mouth and on his tongue,}

{\ \ \ pupils were pinpoint and reactive.}\\

\vspace{-3mm}

{\ \ \ - {\bf{\em blood in his mouth}}}\\
{\ \ \ - {\bf{\em blood \dots on his tongue}}}\\
{\ \ \ - {\bf{\em pupils \dots pinpoint}}}\\
\\
\hline
\\
%
%
%
%
{\ \ \ Infarctions either water shed or embolic } \\

\vspace{-3mm}

{\ \ \ - {\bf{\em Infarctions}}}\\
{\ \ \ - {\bf{\em Infarctions \dots water shed}}}\\
{\ \ \ - {\bf{\em Infarctions \dots embolic}}}\\
\\
\hline
\\
{\ \ \ You see blood or dark/black material when you vomit or}

{\ \ \ have a bowel movement.}\\

\vspace{-3mm}

{\ \ \ - {\bf{\em blood \dots vomit}}}\\

{\ \ \ - {\bf{\em blood \dots bowel movement}}}\\

{\ \ \ - {\bf{\em dark \dots material \dots vomit}}}\\

{\ \ \ - {\bf{\em dark \dots bowel movement}}}\\

{\ \ \ - {\bf{\em black material \dots vomit}}}\\

{\ \ \ - {\bf{\em black material \dots bowel movement}}}\\
\\

\hline
\end{tabularx}
}
\vspace{0mm}
\caption{\footnotesize{Examples of discontiguous and overlapping  mentions, taken from the dataset.}}
\label{ct-sample-2}
\vspace{0mm}
\end{figure}

Figure \ref{ct-sample-2} shows some examples of the mentions. 
The first example shows two discontiguous mentions that do not overlap. The second example shows a typical discontiguous and overlapping case. The last example shows a very hard case of overlapping and discontiguous mentions, as each of the components in \{{\bf{\em blood}}, {\bf{\em dark}}, {\bf{\em black material}}\} is paired with each of the word in \{{\bf{\em vomit}}, {\bf{\em bowel movement}}\}, resulting in six mentions in total, with one having three components (\textbf{\em dark \dots material \dots vomit}).

\begin{table*}[ht!]
\footnotesize
\centering
{\def\arraystretch{1.075}\tabcolsep=4pt
\setlength\doublerulesep{1pt}
\begin{tabular}{l|ccc|ccc||ccc|ccc}
& \multicolumn{6}{c||}{\textsc{small}} & \multicolumn{6}{c}{\textsc{large}}\\
& \multicolumn{3}{c|}{\multirow{1}{*}{Train-Disc}} & \multicolumn{3}{c||}{\multirow{1}{*}{Train-Orig}} & \multicolumn{3}{c|}{\multirow{1}{*}{Train-Disc}} & \multicolumn{3}{c}{\multirow{1}{*}{Train-Orig}}\\\hline
& P & R & F1 & P & R & F1 & P & R & F1 & P & R & F1\\\hline
\multirow{1}{*}{\textsc{Li-enh}} & 59.7  &  39.8  &  47.8  &  71.0  &  45.8  &  55.7 &  54.7  &  41.2  &  47.0 & 64.1 & 46.5  &  53.9\\
\multirow{1}{*}{\textsc{Li-all}} & 16.6  &  \textbf{43.5}  &  24.1 &  55.5  &  \textbf{49.2}  &  52.2  &  15.2  &  \textbf{44.9}  &  22.7 &  52.8  &  49.4  &  51.1 \\\hline
\multirow{1}{*}{\textsc{Sh-enh}} &  85.9  &  39.7  &  \textbf{54.3} &  82.2  &  48.0  &  60.6 &  76.9  &  40.1  &  52.7 & 73.9 & 49.1 &  59.0 \\
\multirow{1}{*}{\textsc{Sh-all}} &  85.9  &  39.7  &  \textbf{54.3}  &  82.2  &  48.0  &  60.6 &  76.0  &  40.5  &  \textbf{52.8} &  73.4  &  \textbf{49.5}  &  59.1\\\hline
\multirow{1}{*}{\textsc{Sp-enh}} &  \textbf{86.7}  &  37.8  &  52.7 &  \textbf{82.5}  &  48.0  &  \textbf{60.7} &  \textbf{79.4}  &  38.6 &  52.0  &  \textbf{75.3}  &  48.8 & \textbf{59.2} \\
\multirow{1}{*}{\textsc{Sp-all}} &  \textbf{86.7}  &  37.8  &  52.7  &  \textbf{82.5}  &  48.0  &  \textbf{60.7} &  \textbf{79.4}  &  38.6  &  52.0  &  \textbf{75.3}  &  48.8  &  \textbf{59.2}
\end{tabular}
}
\vspace{0mm}
\caption{Results on the two datasets and two different training data after optimizing regularization hyperparameter $\lambda$ in development set. The \textsc{-enh} and \textsc{-all} suffixes refer to the \textsc{Enough} and \textsc{All} heuristics. The best result in each column is put in boldface.}
\label{all-result}
\vspace{-1mm}
\end{table*}

\subsection{Features}

Motivated by the features used by Zhang {et al.}~\shortcite{Zhang2014},
for both the linear-chain CRF model and our models we use the following features: neighbouring words  with relative position information (we consider previous and next $k$ words, where $k$=$1,2,3$), neighbouring words with relative position information paired with the current word,  word $n$-grams containing the current word ($n$=2,3), POS tag for the current word,  POS tag $n$-grams containing the current word ($n$=2,3), orthographic features (prefix, suffix, capitalization, lemma), note type (discharge summary, echo report, radiology, and ECG report), section name (\textit{e.g.} Medications, Past Medical History)\footnote{Section names were determined by some heuristics, refer to the supplementary material for more information}, Brown cluster, and word-level semantic category information\footnote{This is standard information that can be extracted from UMLS. See \cite{Zhang2014} for more details.}.
We used Stanford POS tagger \cite{Toutanova2003} for POS tagging, and NLP4J package\footnote{http://www.github.com/emorynlp/nlp4j/} for lemmatization.
For Brown cluster features, following Tang et al. \shortcite{Tang2013}, we used 1,000 clusters from the combination of training, development, and test set, and used all the subpaths of the cluster IDs as features.


\subsection{Experimental Setup}

We evaluated the three models on the \textsc{small} dataset and the \textsc{large} dataset.

Note that in both the \textsc{small} and \textsc{large} dataset, about half of all mentions are discontiguous, both in training and test set. We also want to see whether training on a set where the majority of the mentions are contiguous will affect the performance on recognizing discontiguous mentions. So we also performed another experiment where we trained each model on the original training set where the majority of the entities are contiguous. We refer to this original dataset as ``Train-Orig'' (it contains 10,405 sentences, including those with no entities) and the earlier one as ``Train-Disc''.

First we trained each model on the training set, varying the regularization hyperparameter $\lambda$,\footnote{Taken from the set \{0.125, 0.25, 0.5, 1.0, 2.0\}} then the $\lambda$ with best result in the development set using the respective \textsc{Enough} heuristic for each model is chosen for final result in the test set.


For each experiment setting, we show precision (P), recall (R) and F1 measure. Precision is the percentage of the mentions predicted by the model which are correct, recall is the percentage of mentions in the dataset correctly discovered by the model, and F1 measure is the harmonic mean of precision and recall.


\subsection{Results and Discussions}

The full results are recorded in Table \ref{all-result}.

We see that in general our models have higher precision compared to the linear-chain baseline. This is expected, since our models have less ambiguity, which means that from a given output structure it is easier in our model to get the correct interpretation. We will explore this more in Section \ref{exp:ambiguity}.

The \textsc{All} heuristic, as expected, results in higher recall, and this is more pronounced in the linear-chain model, with up to 4\% increase from the \textsc{Enough} heuristic, achieving the highest recall in three out of four settings. The high recall of the \textsc{All} heuristic in the linear-chain model can be explained by the high level of ambiguity the model has. Since it has more ambiguity compared to our models, one label sequence predicted by the model produces a lot of entities, and so it is more likely to overlap with the gold entities. But this has the drawback of very low precision as we can see in the result.

We see switching from one heuristic to the other does not affect the results of our models much. Looking at the output of our models, they tend to produce output structures with less ambiguity, which causes little difference in the two heuristics.


One example where the baseline made a mistake is the sentence: ``Ethanol Intoxication and withdrawal''. The gold mentions are ``Ethanol Intoxication'' and ``Ethanol withdrawal''. But the linear-chain model labeled it as ``[Ethanol]\subscript{B} [Intoxication]\subscript{I} and [withdrawal]\subscript{BD}'', which is inconsistent since there is only one discontiguous component. Our models do not have this issue because in our models every subgraph that may be predicted translates to valid mention combinations, as discussed in Section \ref{our-model}.

In the ``Train-Orig'' column, we see that all models can recognize discontiguous entities better when given more data, even though the majority of the entities in ``Train-Orig'' are contiguous.

\subsection{Experiments on Ambiguity}
\label{exp:ambiguity}

To see the ambiguity of each model empirically, we run the decoding process for each model given the gold output structure, which is the true label sequence for the linear-chain model and the true mention-encoded hypergraph for our models.

We used the entities from the training and development sets for this experiment, and we compare the ``Original'' datasets with the ``Discontiguous'' subset to see that the ambiguity is more pronounced when there are more discontiguous entities. Then we show the precision and recall errors (defined as $1-P$ and $1-R$, respectively) in Table \ref{encoding-decoding}.

\begin{table}
\centering
\footnotesize
{\def\arraystretch{1.15}\tabcolsep=4pt
\begin{tabular}{l|cc||cc}
\multicolumn{1}{c|}{} & \multicolumn{2}{c||}{Discontiguous} & \multicolumn{2}{c}{Original} \\\cline{2-5}
& Prec Err & Rec Err & Prec Err & Rec Err\\\hline
\textsc{Li-all} & 63.66\% & 0.30\% & 23.81\% & 0.17\% \\
\textsc{Sh-all} & 1.73\% & 0.30\% & 0.35\% & 0.17\% \\
\textsc{Sp-all} & 1.05\% & 0.30\% & 0.22\% & 0.17\% \\\hline
\textsc{Li-enh} & 2.74\% & 3.82\% & 0.52\% & 0.90\% \\
\textsc{Sh-enh} & 1.21\% & 1.46\% & 0.25\% & 0.38\% \\
\textsc{Sp-enh} & 0.75\% & 0.90\% & 0.17\% & 0.28\%
\end{tabular}
}
\vspace{0mm}
\caption{Precision and recall errors (\%) of each model in the ``Discontiguous'' and ``Original'' datasets when given the gold output structure (label sequence in linear-chain model, hypergraph in our models). Lower numbers are better.}
\label{encoding-decoding}
\vspace{-1mm}
\end{table}


\begin{table}
\centering
\footnotesize
{\def\arraystretch{1.15}\tabcolsep=2pt
\begin{tabular}{l|r|ccc|ccc|ccc}
\multirow{2}{*}{Type} & \multirow{2}{*}{\#Ent} & \multicolumn{3}{c|}{Linear-chain} & \multicolumn{3}{c|}{\textsc{Shared}} & \multicolumn{3}{c}{\textsc{Split}} \\
 & & P & R & F & P & R & F & P & R & F\\\hline
\texttt{A} & 289 & 69.8 & 59.9 & 64.4 & 79.4 & 56.1 & 65.7 & 81.0 & 56.1 & 66.3 \\
\texttt{B} & 418 & 50.0 & 34.0 & 40.5 & 56.8 & 29.0 & 38.4 & 58.2 & 28.0 & 37.8 \\
\texttt{N} & 503 & 62.1 & 37.8 & 47.0 & 84.8 & 43.3 & 57.4 & 84.9 & 42.4 & 56.5 \\\hline
Total      &1210 & 60.3 & 41.7 & 49.3 & 74.3 & 41.4 & 53.2 & 75.5 & 40.7 & 52.9
\end{tabular}
}
\vspace{0mm}
\caption{Results on the \textsc{large} dataset when entities are split into three types: \texttt{A}, \texttt{B}, and \texttt{N}. \#Ent is the number of entities.}
\label{multi-type}
\vspace{-1mm}
\end{table}

Since the \textsc{All} heuristics generates all possible mentions from the given encoding, theoretically it should give perfect recall. However, due to errors in the training data, there are mentions which cannot be properly encoded in the models\footnote{There are 19 errors in the original dataset, and 6 in the discontiguous subset, which include duplicate mentions and mentions with incorrect boundaries}. Removing these errors results in perfect recall (0\% recall error). This means that all models are complete: they can encode any mention combinations.

We see however, a very huge difference on the precision error between the linear-chain model and our models, even more when most of the entities are discontiguous.
For the discontiguous subset with the \textsc{All} heuristic, the linear-chain model produced 5,463 entities, while the \Shared and \Split model produced 2,020 and 2,006 entities, respectively. The total number of gold entities is 1,991.
This means one encoding in the linear-chain model produces much more distinct mention combinations compared to our model, which again shows that the linear-chain model has more ambiguity. Similarly, we can deduce that the \Shared model has slightly more ambiguity compared to the \Split model. This confirms our theoretical result presented previously.

It is also worth noting that in the \textsc{Enough} heuristic our models have smaller errors compared to the linear-chain model, showing that when both models can predict the true output structure (the correct label sequence for the baseline model and mention-encoded hypergraph for our models), it is easier in our models to get the desired mention combinations.


\subsection{Experiments on Multiple Entity Types}
We used the \textsc{large} dataset with the multiple-type entities for this experiment. We ran our two models and the linear-chain CRF model with the \textsc{Enough} heuristic on this multi-type dataset, in the same setting as Train-Orig in previous experiments, and the result is shown in Table \ref{multi-type}. We used the best lambda from the main experiment for this experiment.

There is a performance drop compared to the \textsc{large}-Train-Orig results in Table \ref{all-result}, which is expected since the presence of multiple types make the task harder. But in general we still see that our models are still better than the baseline, especially the \textsc{Split} model, which shows that in the presence of multiple types, our models can still work better than the baseline model.

\section{Conclusions and Future Work}

In this paper we proposed new models that can better represent discontiguous entities that can be overlapping at the same time.
We validated our claims through theoretical analysis and empirical analysis on the models' ambiguity, as well as their performances on the task of recognizing disorder mentions on datasets with a substantial number of discontiguous entities.
When the true output structure is given, which is still ambiguous in all models, our models show that it is easier to produce the desired mention combinations compared to the linear-chain CRF model with reasonable heuristics.
We note that an extension similar to semi-Markov or weak semi-Markov \cite{Muis2016} is possible for our models. We leave this for future investigations.

The supplementary material and our implementations for the models are available at:

\url{http://statnlp.org/research/ie}

\section*{Acknowledgments}

We would like to thank the anonymous reviewers for their helpful feedback, and also the ShARe/CLEF eHealth Evaluation Lab for providing us the dataset. This work is supported by MOE Tier 1 grant SUTDT12015008.


\bibliography{emnlp2016-discontiguous-entities}

\begin{thebibliography}{}

\bibitem[\protect\citename{Al-Rfou and Skiena}2012]{Al-Rfou2012}
Rami Al-Rfou and Steven Skiena.
\newblock 2012.
\newblock {SpeedRead: A Fast Named Entity Recognition Pipeline}.
\newblock {\em Proceedings of COLING 2012}, pages 51--66.

\bibitem[\protect\citename{Asahara and Matsumoto}2003]{Asahara2003}
Masayuki Asahara and Yuji Matsumoto.
\newblock 2003.
\newblock {Japanese Named Entity Extraction with Redundant Morphological
  Analysis}.
\newblock In {\em Proceedings of HLT-NAACL '03}, volume~1, pages 8--15.

\bibitem[\protect\citename{Baker}1979]{Baker1979}
James~K Baker.
\newblock 1979.
\newblock {Trainable Grammars for Speech Recognition}.
\newblock {\em Journal of the Acoustical Society of America}, 65(S1):S132.

\bibitem[\protect\citename{Bikel \bgroup et al.\egroup }1997]{Bikel1997}
Daniel~M. Bikel, Scott Miller, Richard~M. Schwartz, and Ralph Weischedel.
\newblock 1997.
\newblock {Nymble: a high-performance learning name-finder}.
\newblock {\em Proceedings of the fifth conference on Applied Natural Language
  Processing (ANLP '97)}, pages 194--201.

\bibitem[\protect\citename{Borthwick and Sterling}1998]{Borthwick1998}
Andrew Borthwick and John Sterling.
\newblock 1998.
\newblock {NYU: Description of the MENE named entity system as used in MUC-7}.
\newblock In {\em Proceedings of the 7th Message Understanding Conference
  (MUC-7)}.

\bibitem[\protect\citename{Finkel and Manning}2009]{Finkel2009}
Jenny~Rose Finkel and Christopher~D. Manning.
\newblock 2009.
\newblock {Nested named entity recognition}.
\newblock In {\em Proceedings of the 2009 Conference on Empirical Methods in
  Natural Language Processing (EMNLP '09)}, volume~1, pages 141--150.

\bibitem[\protect\citename{Florian \bgroup et al.\egroup }2004]{Florian:2004uv}
Radu Florian, Hany Hassan, Abraham Ittycheriah, Hongyan Jing, Nanda Kambhatla,
  Xiaoqiang Luo, H~Nicolov, and Salim Roukos.
\newblock 2004.
\newblock {A statistical model for multilingual entity detection and tracking}.
\newblock In {\em Proceedings of HLT-NAACL '04}, pages 1--8.

\bibitem[\protect\citename{Hasegawa \bgroup et al.\egroup }2004]{Hasegawa2004b}
Takaaki Hasegawa, Satoshi Sekine, and Ralph Grishman.
\newblock 2004.
\newblock {Discovering Relations Among Named Entities from Large Corpora}.
\newblock {\em Proceedings of the 42nd Annual Meeting on Association for
  Computational Linguistics}, pages 415--422.

\bibitem[\protect\citename{Lafferty \bgroup et al.\egroup
  }2001]{Lafferty:2001ty}
John Lafferty, Andrew McCallum, and Fernando~CN Pereira.
\newblock 2001.
\newblock {Conditional random fields: Probabilistic models for segmenting and
  labeling sequence data}.
\newblock In {\em Proceedings of International Conference on Machine Learning
  (ICML '01)}, pages 282--289.

\bibitem[\protect\citename{Lu and Roth}2015]{Lu2015}
Wei Lu and Dan Roth.
\newblock 2015.
\newblock {Joint Mention Extraction and Classification with Mention
  Hypergraphs}.
\newblock In {\em Proceedings of the 2015 Conference on Empirical Methods in
  Natural Language Processing (EMNLP '15)}, pages 857--867.

\bibitem[\protect\citename{McCallum and Li}2003]{McCallum2003}
Andrew McCallum and Wei Li.
\newblock 2003.
\newblock {Early results for named entity recognition with conditional random
  fields, feature induction and web-enhanced lexicons}.
\newblock In {\em Proceedings of HLT-NAACL '03}, volume~4, pages 188--191.

\bibitem[\protect\citename{Muis and Lu}2016]{Muis2016}
Aldrian~Obaja Muis and Wei Lu.
\newblock 2016.
\newblock {Weak Semi-Markov CRFs for Noun Phrase Chunking in Informal Text}.
\newblock In {\em Proceedings of HLT-NAACL '16}, pages 714--719.

\bibitem[\protect\citename{Nadeau and Sekine}2007]{Nadeau2007}
David Nadeau and Satoshi Sekine.
\newblock 2007.
\newblock {A survey of named entity recognition and classification}.
\newblock {\em Lingvisticae Investigationes}, 30(1):3--26.

\bibitem[\protect\citename{Pradhan \bgroup et al.\egroup }2014a]{Pradhan2014}
Sameer Pradhan, No\'{e}mie Elhadad, Wendy~W. Chapman, Suresh Manandhar, and
  Guergana Savova.
\newblock 2014a.
\newblock {SemEval-2014 Task 7: Analysis of Clinical Text}.
\newblock In {\em Proceedings of the 8th International Workshop on Semantic
  Evaluation (SemEval 2014)}, pages 54--62.

\bibitem[\protect\citename{Pradhan \bgroup et al.\egroup }2014b]{Pradhan2014a}
Sameer Pradhan, No\'{e}mie Elhadad, Brett~R. South, David Martinez, Lee
  Christensen, Amy Vogel, Hanna Suominen, Wendy~W. Chapman, and Guergana
  Savova.
\newblock 2014b.
\newblock {Evaluating the state of the art in disorder recognition and
  normalization of the clinical narrative.}
\newblock {\em Journal of the American Medical Informatics Association :
  JAMIA}, 22(1):143--54.

\bibitem[\protect\citename{Ratinov and Roth}2009]{Ratinov2009}
Lev Ratinov and Dan Roth.
\newblock 2009.
\newblock {Design Challenges and Misconceptions in Named Entity Recognition}.
\newblock In {\em Proceedings of the Thirteenth Conference on Computational
  Natural Language Learning (CoNLL '09)}, pages 147--155.

\bibitem[\protect\citename{Sekine}1998]{Sekine1998}
Satoshi Sekine.
\newblock 1998.
\newblock {NYU: Description of the Japanese NE system used for MET-2}.
\newblock In {\em Proceedings of the 7th Message Understanding Conference
  (MUC-7)}.

\bibitem[\protect\citename{Suominen \bgroup et al.\egroup }2013]{Suominen2013}
Hanna Suominen, Sanna Salanter{\"a}, Sumithra Velupillai, Wendy~W. Chapman,
  Guergana Savova, Noemie Elhadad, Sameer Pradhan, Brett~R. South, Danielle~L.
  Mowery, Gareth J.~F. Jones, Johannes Leveling, Liadh Kelly, Lorraine
  Goeuriot, David Martinez, and Guido Zuccon, 2013.
\newblock {\em Overview of the ShARe/CLEF eHealth Evaluation Lab 2013}, pages
  212--231.
\newblock Springer Berlin Heidelberg, Berlin, Heidelberg.

\bibitem[\protect\citename{Tang \bgroup et al.\egroup }2013]{Tang2013}
Buzhou Tang, Yonghui Wu, Min Jiang, Joshua~C. Denny, and Hua Xu.
\newblock 2013.
\newblock {Recognizing and Encoding Discorder Concepts in Clinical Text using
  Machine Learning and Vector Space}.
\newblock In {\em Proceedings of the ShARe/CLEF Evaluation Lab}.

\bibitem[\protect\citename{Toutanova \bgroup et al.\egroup
  }2003]{Toutanova2003}
Kristina Toutanova, Dan Klein, and Christopher~D Manning.
\newblock 2003.
\newblock {Feature-rich part-of-speech tagging with a cyclic dependency
  network}.
\newblock In {\em Proceedings of HLT-NAACL '03}, volume~1, pages 252--259.

\bibitem[\protect\citename{Xu \bgroup et al.\egroup }2015]{Xu2015}
Jun Xu, Yaoyun Zhang, Jingqi Wang, Yonghui Wu, and Min Jiang.
\newblock 2015.
\newblock {UTH-CCB : The Participation of the SemEval 2015 Challenge – Task
  14}.
\newblock In {\em Proceedings of the 9th International Workshop on Semantic
  Evaluation (SemEval 2015)}, pages 311--314.

\bibitem[\protect\citename{Zhang \bgroup et al.\egroup }2014]{Zhang2014}
Yaoyun Zhang, Jingqi Wang, Buzhou Tang, Yonghui Wu, Min Jiang, Yukun Chen, and
  Hua Xu.
\newblock 2014.
\newblock {UTH\_CCB: A report for SemEval 2014 -- Task 7 Analysis of Clinical
  Text}.
\newblock In {\em Proceedings of the 8th International Workshop on Semantic
  Evaluation (SemEval 2014)}, pages 802--806.

\end{thebibliography}


\begin{thebibliography}{}

\bibitem[Lu and Roth, 2015]{Lu2015}
Lu, W. and Roth, D. (2015).
\newblock {Joint Mention Extraction and Classification with Mention
  Hypergraphs}.
\newblock In {\em Proceedings of the 2015 Conference on Empirical Methods in
  Natural Language Processing (EMNLP'15)}, pages 857--867.

\bibitem[Muis and Lu, 2016]{Muis2016b}
Muis, A.~O. and Lu, W. (2016).
\newblock {Learning to Recognize Discontiguous Entities}.
\newblock In {\em Proceedings of the 2016 Conference on Empirical Methods in
  Natural Language Processing (EMNLP '16)}.

\bibitem[Tang et~al., 2013]{Tang2013}
Tang, B., Wu, Y., Jiang, M., Denny, J.~C., and Xu, H. (2013).
\newblock {Recognizing and Encoding Discorder Concepts in Clinical Text using
  Machine Learning and Vector Space}.
\newblock In {\em Proceedings of the ShARe/CLEF Evaluation Lab}.

\end{thebibliography}
\bibliographystyle{emnlp2016}

\end{document}


\maketitle
\begin{abstract}
This is the supplementary material for ``Learning to Recognize Discontiguous Entities'' \cite{Muis2016b}. This material gives more details in the experiment setup, the ambiguity of each model, and compare the models from theoretical point of view.
\end{abstract}

\section{Model Ambiguity}

This work attempts to define a more formal way to compare two models in terms of its ambiguity.

To compare the ambiguity between models, we first define a notion of \textit{model ambiguity level}, which is defined as the average number of distinct interpretations across its set of \textit{canonical encodings}. 
A canonical encoding is a fixed, selected representation of a particular set of mentions in a sentence, among (possibly) several alternative representations. Several alternatives may be present due to the ambiguity of the encoding-decoding process and also since the output of the model is not restricted to a specific rule. For example, for the text ``John Smith'', a model trained in BIO format might output ``B-PER I-PER'' or ``I-PER I-PER'', and both will still mean that ``John Smith'' is a person, although the ``correct'' encoding would of course be ``B-PER I-PER'', which is selected as the canonical encoding. Intuitively, a canonical encoding is a formal way to say that we only consider the ``correct'' encodings.

Note that by definition, the number of canonical encodings cannot be larger than the number of possible interpretations, so ambiguity level will be at least 1. When the number of canonical encodings is strictly smaller than the number of possible entity combinations, we have either ambiguity (when several entity combinations have the same canonical encoding) or incompleteness (when a entity combination does not have an encoding). A model which is complete and not ambiguous will have the lowest ambiguity level, which is 1. We note that in our case, all models are complete as for any entity combination each model is able to encode it. All models that we consider, however, are ambiguous since some in each model there are encodings that can be interpreted in more than one way.

Let $\mathcal{N}(n)$ be the number of all possible entity combinations in a sentence with $n$ words, and $\mathcal{M}(n)$ be the number of canonical encodings in the model for a sentence with $n$ words. Then the ambiguity level of a model is:

\vspace{-5mm}

\begin{equation}
A = \lim_{n\to\infty}\frac{\sum_{i=1}^n \mathcal{N}(i)}{\sum_{i=1}^n\mathcal{M}(i)}
\end{equation}

\noindent assuming the limit exists. If it does not exist, the ambiguity is undefined. This can mean two models which are incomparable, or that the number of possible entity combinations is much larger than the number of canonical encodings. When it exists, the values of $A$ lie in the range $[1,\infty)$. Sometimes when $\mathcal{M}(n)$ is very small compared to $\mathcal{N}(n)$, it might be useful to compute the \textit{log-ambiguity} instead, defined as:
\begin{equation}
\mathcal{A} = \lim_{n\to\infty}\frac{\log\left(\sum_{i=1}^n \mathcal{N}(i)\right)}{\log\left(\sum_{i=1}^n\mathcal{M}(i)\right)}
\end{equation}

This is what we will use in the remainder of this article.

\paragraph{Relative Ambiguity} For some tasks, the number of possible encodings in a practical model may be very small compared to the number of possible entity combinations resulting in very high ambiguity, and so it is difficult to compare two models. To overcome this, we define a notion of \textit{relative ambiguity}, which is defined as the ratio of the ambiguity level of two models:
\begin{equation}
\mathcal{A}_r(M_1, M_2) = \frac{\mathcal{A}_{M_1}(n)}{\mathcal{A}_{M_2}(n)} = \lim_{n\to\infty} \frac{\log\sum_{i=1}^n\mathcal{M}_{M_2}(i)}{\log\sum_{i=1}^n\mathcal{M}_{M_1}(i)}
\end{equation}

A relative ambiguity greater than one means the first model is more ambiguous compared to the second model. A relative ambiguity of 1 means the two models have the same level of ambiguity.


Now we calculate the number of canonical encodings $\mathcal{M}_{\textsc{Li}}(n)$, $\mathcal{M}_{\textsc{Sh}}(n)$, and $\mathcal{M}_{\textsc{Sp}}(n)$ for linear-chain, \textsc{Shared}, and \textsc{Split} model, respectively.

\paragraph{Calculating $\mathcal{M}_{\textsc{Li}}(n)$} For the linear-chain, the number of possible encodings is the number of possible tag sequence, which is $7^n$ in this case, since there are 7 possible tags. Note that this number is larger than the number of canonical encodings, since some of the tag sequences have the same interpretation and some are not valid. So: $\sum_{i=1}^n \mathcal{M}_{\textsc{Li}}(i) < \sum_{i=1}^n 7^i = \frac{7^{n+1}-7}{6} < 8^n < 2^{3n}$ and so $\log \sum_{i=1}^n\mathcal{M}_{\textsc{Li}}(i) < 3n\log 2$.

\paragraph{Calculating $\mathcal{M}_{\textsc{Sh}}(n)$ and $\mathcal{M}_{\textsc{Sp}}(n)$}For our models, the number of canonical encodings is equal to the number of valid subgraph in the model, since each distinct subgraph will yield distinct interpretation. This is quite tricky to calculate; a straightforward application of the standard dynamic programming algorithm similar to the inside algorithm that uses the nodes as states fails in this case because it also includes certain combinations not representable in our graph models. To calculate the number of distinct subgraph, we need to use a combination of nodes as states.

\begin{figure}[h]
\centering
\begin{minipage}{0.4\textwidth}
\includegraphics[width=\textwidth]{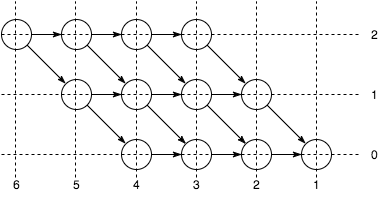}
\vspace{-15pt}
\caption{A simple graph.}
\label{grid-simple}
\vspace{-5pt}
\end{minipage}
\end{figure}

To explain this idea, we will calculate the number of subgraph of a simple graph shown in Figure \ref{grid-simple}.The nodes are indexed by position from right to left, starting from 1, and also by level from bottom to top, starting from 0, as in the figure. So the top left node has level 2 and position 6, and the bottom right node has level 0 and position 1. Let $n_i^j$ denote the node at level $i$ and position $j$. From each node $n_i^j$ except the nodes in the bottom row and the rightmost in each level, there are two edges, one connecting to $n_i^{j-1}$ and another connecting to $n_{i-1}^{j-1}$. Now we want to count the number of distinct connected directed acyclic graphs (DAGs) that include the top left node and the bottom right node.

First note that at each position, any combination of nodes in the three levels can be assigned a unique number from 0 to 7 by considering the chosen nodes as binary number, with the node at level 2 being the most significant bit. For example, the number 5, or $\overline{101}$ in binary, represents the set of two nodes at level 2 and level 0.

\begin{figure}[t]
\centering
\begin{minipage}{0.9\textwidth}
\includegraphics[width=\textwidth]{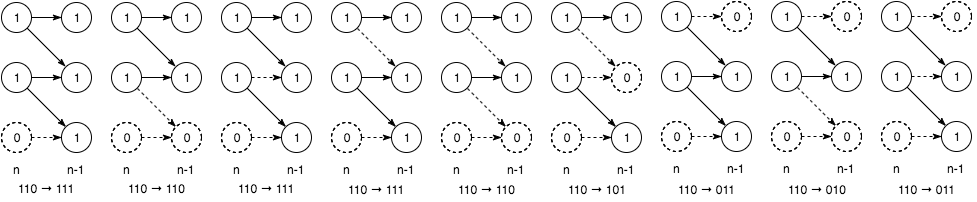}
\vspace{-15pt}
\caption{Reachable node states from the state $\overline{110}$.}
\label{node-states}
\vspace{-5pt}
\end{minipage}
\vspace{-3mm}
\end{figure}

Now, we define 8 functions, $f_{\overline{000}}(n)$, $f_{\overline{001}}(n)$, $\ldots$, $f_{\overline{111}}(n)$, one for each combination of nodes at each position. The function $f_{k}(n)$ represents the number of directed acyclic graphs with the node combinations represented by $k$ at position $n$ as the sources, and the bottom right node as the sink. Then we have each function as the sum over all reachable node combinations at the previous position. To find the reachable states, we enumerate all possible edge combinations for a given nodes configuration. Figure \ref{node-states} shows the 9 reachable states from the state $\overline{110}$. Note that a node combination may be reachable in multiple ways.

From there, we have $f_{\overline{110}}(n) = f_{\overline{010}}(n-1) + 2f_{\overline{011}}(n-1) + f_{\overline{101}}(n-1) + 2f_{\overline{110}}(n-1) + 3f_{\overline{111}}(n-1)$. By representing this as transition vector, we have:

\vspace{-5mm}

\begin{equation}
f_{\overline{110}}(n) = {\begin{bmatrix}0&0&1&2&0&1&2&3\end{bmatrix}} \times
\begin{bmatrix}
f_{\overline{000}}(n-1)\\
f_{\overline{001}}(n-1)\\
f_{\overline{010}}(n-1)\\
f_{\overline{011}}(n-1)\\
f_{\overline{100}}(n-1)\\
f_{\overline{101}}(n-1)\\
f_{\overline{110}}(n-1)\\
f_{\overline{111}}(n-1)
\end{bmatrix} = \begin{bmatrix}0&0&1&2&0&1&2&3\end{bmatrix}\times \mathbf{f}(n-1)\nonumber
\end{equation}

Stacking the transition vectors for all 8 functions, we have the following transition matrix $\mathbf{T}$:

\vspace{-3mm}

\begin{equation}
\mathbf{T} = 
	\begin{bmatrix}
	0&0&0&0&0&0&0&0\\
	0&1&0&0&0&0&0&0\\
	0&1&1&1&0&0&0&0\\
	0&1&0&2&0&0&0&0\\
	0&0&1&0&1&0&1&0\\
	0&0&0&1&0&1&0&1\\
	0&0&1&2&0&1&2&3\\
	0&0&0&3&0&1&0&5\\
	\end{bmatrix}
	\quad
	\text{, and }
	\qquad
	\mathbf{f}(n) = \mathbf{T}\cdot\mathbf{f}(n-1)
	\nonumber
\end{equation}

Using this recursive formulation, we can calculate the number of possible DAGs at the top left node by calculating $f_{\overline{100}}(n)$. Initially we have $f_{\overline{001}}(1)=1$ and $f_k(1) = 0$ for $k \neq 1$. Then we have the following formulation for $f(n)$, the number of DAGs from the top left node at position $n\geq3$ to the bottom right node:

\begin{equation}
f(n) = {\begin{bmatrix}0&0&0&0&1&0&0&0\end{bmatrix}} \times \mathbf{T}^{n-1} \times {\begin{bmatrix}0&1&0&0&0&0&0&0\end{bmatrix}}^T
\nonumber
\end{equation}

In general, when a transition matrix is diagonalizable into $\mathbf{T} = \mathbf{S}^{-1}\mathbf{J} \mathbf{S}$ for some matrix $\mathbf{S}$ and diagonal matrix $\mathbf{J}$, the value of $\mathbf{T}^{n}$ can be easily calculated by exponentiating the diagonal entries in $\mathbf{J}$ since $\mathbf{T}^n = \mathbf{S}^{-1}\mathbf{J}^n\mathbf{S}$. Note that when $\mathbf{T}$ is diagonalizable, the entries in $\mathbf{J}$ will be the eigenvalues of $\mathbf{T}$ and there will be coefficients $c_{ij}$ such that:

\begin{equation}
f(n) = \sum_{i=1}^k \left(\sum_{j=1}^{q_i} c_{ij}\cdot n^{j-1}\cdot{\lambda_i}^n\right)
\end{equation}

\noindent where $\lambda_1, \lambda_2, \ldots, \lambda_k$ are the eigenvalues of $\mathbf{T}$ and $q_i$ is the multiplicity of $\lambda_i$.

Suppose $\lambda_m = \max(\lambda_1, \lambda_2, \ldots, \lambda_k)$, then $n^{q_m}{\lambda_m}^n$ will be the dominating term, and the value of $f(n)$ will asymptotically grow as fast as $n^{q_m}{\lambda_m}^n$.

Equivalently, $f(n) \in \Theta(n^{q_m}\cdot{\lambda_m}^n)$, where:
\begin{eqnarray}
f(n) \in \Theta(g(n)) & \Leftrightarrow & f(n) \in \Omega(g(n)) \wedge f(n) \in O(g(n))\\
f(n) \in \Omega(g(n)) & \Leftrightarrow & \exists k_1,n_1, \forall n>n_1, f(n) \geq k_1\cdot g(n)\\
f(n) \in O(g(n)) & \Leftrightarrow & \exists k_2,n_2, \forall n>n_2, f(n) \leq k_2\cdot g(n)
\end{eqnarray}

Note that if $f(n) \in \Theta(g(n))$ and $h(n) < g(n), \forall n$, then $f(n) \in \Omega(h(n))$. Similarly, if $f(n) \in \Theta(g(n))$ and $h(n) > g(n), \forall n$, then $f(n) \in O(h(n))$.

For example, the matrix $\mathbf{T}$ above has the eigenvalues: $3 + \sqrt{5}, 3-\sqrt{5}, 2, 1$ with $2$ and $1$ having the multiplicity of 2 and 4, respectively. And after solving for the $c_{ij}$ we have:

%
%

\begin{equation}
f(n) = \frac{(3+\sqrt{5})^{n-1}}{4\sqrt{5}+10} - \frac{(3-\sqrt{5})^{n-1}}{4\sqrt{5}-10} - 1
\label{char-simple}
\end{equation}

From Equation \ref{char-simple}, we see that the function $f(n)$ grows with order $(3+\sqrt{5}) \approx 5.24$.

Now, applying the same method to our original hypergraph, we can get the order in which the number of canonical encodings grows. For the $\textsc{Shared}$ model with at most two components, we have the following transition matrix:\footnote{The code to calculate the matrix is available at \url{https://github.com/justhalf/discontiguous_entities_emnlp2016/blob/master/count_graph.py}}

\begin{equation}
\mathbf{T_{\textsc{Shared}_2}} = 
	\begin{bmatrix}
	1&0&0&0&1&0&0&0\\
	1&2&0&0&1&2&0&0\\
	0&1&1&1&0&1&1&1\\
	0&3&1&5&0&3&1&5\\
	1&0&2&0&5&0&6&0\\
	1&2&2&4&5&10&6&12\\
	0&1&3&5&0&5&11&17\\
	0&3&3&21&0&15&11&73\\
	\end{bmatrix} \nonumber
\end{equation}

\noindent with the maximum eigenvalue $\lambda_m \approx 80.61$ with multiplicity of 1. Therefore the number of possible DAGs in the \textsc{Shared} model with two components, which is the number of canonical encodings, is in the order of $\Omega(80^n)$. Analogously, we have the largest eigenvalue for the transition matrix for the \textsc{Shared} model with three components to be: $\lambda_m \approx 1261.86 > 1024 = 2^{10}$ with multiplicity of 1, and so $\mathcal{M}_{\textsc{Sh}}(n) \in \Omega(2^{10n})$.

Now, by the definition of $\Omega$, for some $n_0$, we have for all $n>n_0$: $\mathcal{M}_{\textsc{Sh}}(n) \geq C\cdot 2^{10n}$ for some constant $C>0$ and so $\forall n>n_0, \log\sum_{i=1}^n\mathcal{M}_{\textsc{Sh}}(i) > \log \mathcal{M}_{\textsc{Sh}}(n) \geq \log C + \log 2^{10n} = \log C + 10n\log 2$.

Now we can calculate the relative ambiguity between the linear-chain model and our \textsc{Shared} model as:

\begin{equation}
\mathcal{A}_r(\textsc{Li},\textsc{Sh}) = \lim_{n\to\infty} \frac{\log\sum_{i=1}^n\mathcal{M}_{\textsc{Sh}}(i)}{\log\sum_{i=1}^n\mathcal{M}_{\textsc{Li}}(i)} \geq \lim_{n\to\infty} \frac{\log C + 10n\log 2}{3n\log 2} = \frac{10}{3} > 1
\end{equation}

This means the linear-chain model is more ambiguous compared to our model.

Similarly, we can compute the growth order of our \textsc{Split} model to arrive at the conclusion that $\mathcal{M}_{\textsc{Sp}} \in \Omega(2^{15n})$. Since $\mathcal{M}_{\textsc{Sh}}(n) \in O(2^{11n})$, we have $\mathcal{A}_r(\textsc{Sh}, \textsc{Sp}) > \frac{15}{11} > 1$.

\paragraph{Number of canonical encodings for small $n$} One may be concerned that since the analysis above concerns asymptotic values, it is applicable only for large $n$. To address that concern, we show in Table \ref{value-for-small-n} the number of canonical encodings in the linear-chain model and our models, compared with the number of possible entity combinations. Note that it is hard to calculate the number of canonical encodings for the linear-chain model, since the labels interact in a complex way. For smaller $n$, we can exhaustively enumerate the possible encodings, but as $n$ increases it becomes not feasible, so for $n \geq 4$ we show the upperbound instead, which is $7^n$. We will show how to calculate $\mathcal{N}(n)$, the number of all possible entity combinations, later.

\begin{table}[h]
\centering
$\begin{array}{c||c|c|c||c}
n & \mathcal{M}_{\textsc{Li}}(n) & \mathcal{M}_{\textsc{Sh}}(n) & \mathcal{M}_{\textsc{Sp}}(n) & \mathcal{N}(n)\\\hline
1 & 2 & 2 & 2 & 2^1 = 2 \\
2 & 8 & 8 & 8 & 2^3 = 8 \\
3 & 46 & 80 & 80 & 2^7 = 128 \\
4 & < 2401 & 3584 & 6656 & 2^{15} = 32768 \\
5 & < 16807 & 533504 & 2367488 & 2^{31} = 2147483648
\end{array}$
\caption{The number of possible encodings for small values of $n$.}
\label{value-for-small-n}
\end{table}

For $n=1,2$, the number of possible entity combinations is small, and we see that all models can accurately represents each of them. However, note that our models have the advantage of not having the possibility to output non-canonical encodings for $n=1,2$, unlike the linear-chain model, which has 7 and 49 possible encodings for $n=1,2$, respectively. For $n=3$, exhaustive enumeration of the encodings for the linear-chain model shows that there are 46 canonical encodings, which is less than the 80 canonical encodings in our models. For $n \geq 4$, which is typical for sentence length, we see that our models have more canonical encodings compared to the linear-chain model, which is bounded above by $7^n$, although they are still smaller than the total number of possible entity combinations.

\paragraph{Calculating $\mathcal{N}(n)$} Now we show how to calculate the number of possible entity combinations in a sentence of length $n$. In the paper \cite{Lu2015}, the authors had established that the number of possible entity combinations when there is no discontiguous entity is $2^{t\frac{n(n+1)}{2}}$, where $t$ is the number of possible entity types. For this case, we set $t=1$ as there is only one entity type in this task and use the combinatorial identity $\binom{n+1}{2}=\frac{n(n+1)}{2}$ to have $2^{\raisebox{6px}{$\binom{n+1}{2}$}}$ as the number of contiguous entities.

For the number of discontiguous entities with exactly $k$ components, there is one start position and one end position for each of the $k$ components, and all of them are distinct positions. So the number of discontiguous entities is the number of ways to choose the $2k$ positions from $n+1$ positions, resulting in $\binom{n+1}{2k}$ possible discontiguous entities with exactly $k$ components. The total number of discontiguous entities with at most $k$ components is then $\displaystyle N = \sum_{i=1}^{k} \binom{n+1}{2i}$. Theoretically, each of these entities can exist independently of each other, resulting in possibly overlapping entities. Considering the overlapping entities, in total we have:
\begin{equation}
\displaystyle\mathcal{N} = 2^{N}
\end{equation}
\noindent distinct entity combinations with each discontiguous entities having at most $k$ components.

Note that as $k$ goes to $\frac{n+1}{2}$ (which means when we do not restrict the number of components an entity can have), the formula above reduces to $2^{2^n-1}$ by the combinatorial identity $\displaystyle\sum_{i=0}^{\left\lfloor\frac{n+1}{2}\right\rfloor} \binom{n+1}{2i} = 2^n$, which matches the interpretation that when we do not restrict the number of components, then any non-empty subset of words can be an entity, for which there are $2^n-1$ of them. Then since we count entity combinations, we have $2^{2^n-1}$ combinations, which is what we got when we let $k$ goes to $\frac{n+1}{2}$.

In our case, since we are calculating the number of all possible entity combinations for $n\leq 5$, the maximum number of components an entity can have is 3, so the number of all possible entity combinations is also the number of entity combinations with at most 3 components, which is also what we calculated for the models.

\section{Data Preprocessing}
We performed some preprocessing on the clinical texts, which are: sentence splitting, tokenization, and POS tagging.

\paragraph{Splitting and tokenization} We used the document preprocessor from Stanford CoreNLP package to split the documents into sentences, then further processed the output using some rules to better capture the structure of the document. We then tokenized each sentence using a regex-based tokenization, similar to NLTK \texttt{wordpunct\_tokenize} function. We again further processed the output to handle special anonymization tokens (\textit{e.g.}, ``[**doctor first name 77**]'') and normalized them into categories (\textit{e.g.}, ``doctor\_name''). After tokenization, we ran Stanford POS tagger on the resulting text to get the POS tags for each token.
Finally, we assigned each disorder mention into the corresponding sentence that contains it.

Note that in this process there might be disorder mentions which do not fit into any sentence. This happens when the sentence splitter split two words that are annotated as part of a single entity. In our case, we found that there is only one entity in the training set which is incorrectly annotated to also include the whitespace preceeding a sentence, which is not part of any sentence after sentence splitting. We fixed the annotation by excluding the preceding whitespace.

\paragraph{Determining section names} The clinical texts in the datasets are semi-structured, in the sense that the contents are organized into sections. However, the section names seem to be quite irregular, having quite a number of variants of the section names with the same meaning. For the purpose of determining the section name during feature extraction, we used a regular expression to capture lines in the clinical texts that end with a colon or coming after a new line, with some heuristics to handle special cases found in the datasets. More details can be read in our code at \url{http://statnlp.org/research/ie/\#discontiguous-mention}

\section{Handling Ambiguity}

Both the linear-chain CRF model and our models are still ambiguous to some degree, so during the decoding process we need to handle the ambiguity to produce the mentions.
For all models, we define two general heuristics: \textsc{Enough} and \textsc{All}. The \textsc{Enough} heuristic handles ambiguity by trying to produce a minimal set of entities which encodes to the one produced by the model, while \textsc{All} heuristic handles ambiguity by producing the union of all possible entity combinations that encode to the one produced by the model.

More specifically, for the linear-chain CRF baseline, we first converted the labeled words into a sequence of component types: \textit{contiguous}, \textit{body}, and \textit{head}, referring to the sequence of words encoded by \{\textbf{B, I}\}, \{\textbf{BD, ID}\}, and \{\textbf{BH, IH}\}, respectively.
In the \textsc{All} heuristic we form all possible combinations between the components, considering the compatibility between components (\textit{e.g.}, a \textit{body} component cannot be paired with a \textit{contiguous} component).

In the \textsc{Enough} heuristic, each \textit{head} component forms at most two entities by pairing it with \textit{body} components starting from the closest ones from its left, unless there is only one remaining \textit{body} component of that type after this process, in which case the \textit{head} component will form the third entity by being paired with this last \textit{body} component. After this, each unpaired \textit{body} components of the same type will be paired up to form entities with two or three components.
In \cite{Tang2013} the authors mentioned that they used a few simple rules to convert labels to entities, but the exact details on the rules were not made available.

For our models, for \textsc{All}, we generate all possible sets of mentions encoded in the entity-encoded hypergraph by traversing all possible paths in the hypergraph, while in \textsc{Enough}, we generate as many mentions as required to cover all edges present in the hypergraph.

\section{Regularization Hyperparameter}

For the experiments, the regularization hyperparameters for each setting are noted in Table \ref{reg-param}. Note that the same regularization hyperparameter is used for both the \textsc{All} and \textsc{Enough} heuristics during testing.

\begin{table}[h]
\centering
{\def\arraystretch{01}\tabcolsep=4pt
\begin{tabular}{l|cc|cc}
& \multicolumn{2}{c|}{\textsc{small}} & \multicolumn{2}{c}{\textsc{large}}\\
& Train-Disc & Train-Orig & Train-Disc & Train-Orig \\\hline
\textsc{Li} & 0.25 & 0.125 & 0.125 & 0.25 \\
\textsc{Sh} & 0.5 & 0.125 & 0.125 & 0.125 \\
\textsc{Sp} & 2.0 & 0.125 & 0.5 & 0.25
\end{tabular}
}
\caption{Regularization hyperparameter $\lambda$ for each setting.}
\label{reg-param}
\end{table}

\section{Splitting the Dataset}

We note that the dataset that we have only contains one type of entity. In order to evaluate the models on multiple types, we create another dataset where we split the entities based on the entity-level semantic category.

Each entity in the dataset was annotated either with its Concept Unique Identifier (CUI), or with the string ``CUI-less''. The CUI is a number referencing certain entry in the Unified Medical Language System (UMLS) Metathesaurus. In UMLS, each entry has a corresponding semantic type, which is organized in a hierarchy. There are two major roots in the hierarchy, which are type \texttt{A} and type \texttt{B}. These two types, together with the CUI-less entities, make up the three entity types that we used in creating the dataset with multiple types.

\section{Efficiency in Handling Multiple Entity Types}

Theoretically, our models can handle multiple entity types more efficiently compared to the linear-chain CRF model. This is because the time complexity of our models are linear in terms of number of entity types, while it is quadratic for the linear-chain CRF model.

To see this empirically, we randomly split the entities in the dataset into $n$ types, where $n=1,2,4,8,16$, and took note of the time taken for training the models. We show the time per iteration relative to the time taken for handling only one type in Table \ref{multi-type-timing}. We can see that the time per training iteration in the baseline model increases faster than that of our model's, confirming the theoretical time complexity of the models.

\begin{table}[h!]
\centering
\begin{tabular}{l|rrrrr}
\multicolumn{1}{r|}{\#Types} & 1 & 2 & 4 & 8 & 16 \\\hline
Linear CRF & 1.0 & 2.015 & 5.018 & 12.174 & 35.313 \\
\textsc{Shared} & 1.0 & 2.067 & 4.552 & 9.455 & 19.272
\end{tabular}
\caption{The time taken per training iteration for the baseline linear-chain CRF model (Linear CRF) and our \textsc{Shared} model, relative to the time taken for handling one type.}
\label{multi-type-timing}
\end{table}

\bibliography{supp}
\bibliographystyle{apalike}